\PassOptionsToPackage{doi=false,isbn=false,url=false}{bibtex}
\documentclass[letterpaper, 10 pt, conference]{ieeeconf} 

\IEEEoverridecommandlockouts                              

\usepackage{graphicx} 
\graphicspath{ {./figures/} }
\usepackage{amsmath} 
\usepackage{amssymb}
\usepackage{algorithm}
\usepackage[noend]{algpseudocode}
\usepackage{xcolor}
\definecolor{tumblue}{RGB}{0,101,189}

\usepackage{stfloats}
\usepackage{setspace}
\usepackage{tabularx}
\usepackage{booktabs}
\usepackage[T1]{fontenc}
\usepackage[scaled=0.85]{beramono}
\usepackage{listings}
\usepackage{caption}
\usepackage{cleveref}
\usepackage{url}

\Crefname{lstlisting}{Listing}{Listings}
\usepackage{listings}
\usepackage{multirow}
\lstdefinestyle{sparqlstyle}{
  language=OCL,
  basicstyle=\footnotesize,
  stepnumber=1,
  numbersep=10pt,
  tabsize=2,
  showspaces=false,
  breaklines=true
}

\usepackage[nonumberlist,nopostdot,acronym,toc]{glossaries}
\newacronym{bt}{BT}{Behavior Tree}
\newacronym{fsm}{FSM}{Finite State Machine}
\newacronym{llm}{LLM}{Large Language Model}
\newacronym{6dof}{6DOF}{Six Degrees of Freedom}
\newacronym{hrc}{HRC}{Human-Robot Collaboration}
\newacronym{asr}{ASR}{Automatic Speech Recognition}
\newacronym{hra}{HRA}{Human Resource Agent}

\usepackage{balance} 

\makeatletter
\newcommand\fs@betterruled{%
  \def\@fs@cfont{\bfseries}\let\@fs@capt\floatc@ruled
  \def\@fs@pre{\vspace*{5pt}\hrule height.8pt depth0pt \kern2pt}%
  \def\@fs@post{\kern2pt\hrule\relax}%
  \def\@fs@mid{\kern2pt\hrule\kern2pt}%
  \let\@fs@iftopcapt\iftrue}
\floatstyle{betterruled}
\restylefloat{algorithm}
\makeatother

\author{Yifei Li$^{1}$$^{*}$, Jeongwon Park$^{2}$, Guha Manogharan$^{1}$, Feng Ju$^{2}$, and Ilya Kovalenko$^{1}$
\thanks{$^{1}$Pennsylvania State University, University Park, PA, USA
        }%
\thanks{$^{2}$Arizona State University, Tempe, AZ, USA
        }%
\thanks{$^{*}$Corresponding author.}
}

\title{\LARGE \bf
A Mobile Additive Manufacturing Robot Framework for Smart Manufacturing Systems
}

\begin{document}

\setlength{\textfloatsep}{5pt}
\maketitle
\thispagestyle{empty}
\pagestyle{empty}

\begin{abstract}
Recent technological innovations in the areas of additive manufacturing and collaborative robotics have paved the way toward realizing the concept of on-demand, personalized production on the shop floor.
Additive manufacturing process can provide the capability of printing highly customized parts based on various customer requirements.
Autonomous, mobile systems provide the flexibility to move custom parts around the shop floor to various manufacturing operations, as needed by product requirements.
In this work, we proposed a mobile additive manufacturing robot framework for merging an additive manufacturing process system with an autonomous mobile base.
Two case studies showcase the potential benefits of the proposed mobile additive manufacturing framework.
The first case study overviews the effect that a mobile system can have on a fused deposition modeling process. 
The second case study showcases how integrating a mobile additive manufacturing machine can improve the throughput of the manufacturing system.
The major findings of this study are that the proposed mobile robotic AM has increased throughput by taking advantage of the travel time between operations/processing sites.
It is particularly suited to perform intermittent operations (e.g., preparing feedstock) during the travel time of the robotic AM.
One major implication of this study is its application in manufacturing structural components (e.g., concrete construction, and feedstock preparation during reconnaissance missions) in remote or extreme terrains with on-site or on-demand feedstocks.
\end{abstract}


\section{Introduction}
\label{sec:intro}
Rapid advances in manufacturing technology have the potential to significantly increase the productivity, quality, and efficiency of manufacturing systems.
Factories of the future need to have greater personalization capabilities, with each item made to order, at the low cost and high-quality consumers have come to expect~\cite{hu_evolving_2013,lu2017smart}.
To achieve this objective, it is important to develop flexible and adaptable manufacturing systems that produce customized products on-demand~\cite{kovalenko_toward_2022,kovalenko_opportunities_2023-1}.
Two technological enablers that have been proposed to greatly improve various types of flexibility for manufacturing systems are Additive Manufacturing (AM)~\cite{eyers_flexibility_2018} and mobile robotics \cite{fragapane_increasing_2022}.
The AM process consists of creating a product in a layer-by-layer fashion.
The nature of the AM process allows for the customization of individual parts between sequential runs of the AM machine~\cite{pham_additive_2011} or even between layers~\cite{guidetti_data-driven_2022}.
In the past several years, there has been a sharp rise of interest in Industrial AM technology from manufacturing companies~\cite{bromberger_mainstreaming_2022}.
Furthermore, Robotic AM has recently been proposed to allow for greater part customization and to enable the building of more complex parts~\cite{fry_robotic_2020,zhang_robotic_2015}.
Similarly, the use of autonomous mobile robot (AMR) technology has been increasing in factories~\cite{vavra_mobile_2022}.
However, manufacturers primarily use AMRs to create more flexible logistic systems via material handling applications.

Recently, mobile additive manufacturing technology has been proposed to combine Robotic AM and AMR technology.
For example, mobile AM technology has been proposed to create large structures for construction applications~\cite{dorfler_additive_2022}.
For this application, the robotic system creates a large structure in a layer-by-layer manner, moving around the structure to deposit material when necessary.
However, the proposed approach might not be applicable for smaller parts, such as the various components required in various manufacturing systems.
In addition, these approaches do not “manufacture and deliver” a required product, as the produced part is fixed to one location. In ~\cite{diri_kenger_integrated_2021}, the “manufacture and deliver” approach was considered as a potential way to improve the scheduling and planning process of manufacturing processes. However, the scheduling problem assumed that the mobile manufacturing system is a vehicle with a 3D printer and the manuscript does not provide information regarding the construction or control of the robotic system.

In this work, we propose a generic AM and AMR robotic framework that can be used to “manufacture and deliver” components in smart manufacturing systems. We have identified the need to study the development of the system at various levels of manufacturing control system. At the process level, it is necessary to develop an extrusion attachment that leverages existing robotic additive manufacturing technology to rapidly manufacture parts. At the machine level, the robot needs to consider the manufacturing environment and should adapt the movement of the extruder and the AMR to consider dynamic changes in the manufacturing environment. At the system level, there is a need to integrate more advanced scheduling and planning methods that consider the increased flexibility, capabilities, and complexity of the proposed system. The proposed system has the potential to revolutionize existing manufacturing systems through enhanced mobile and additive manufacturing capabilities.

\begin{figure}[t]
\centering
\includegraphics[width=1\columnwidth]{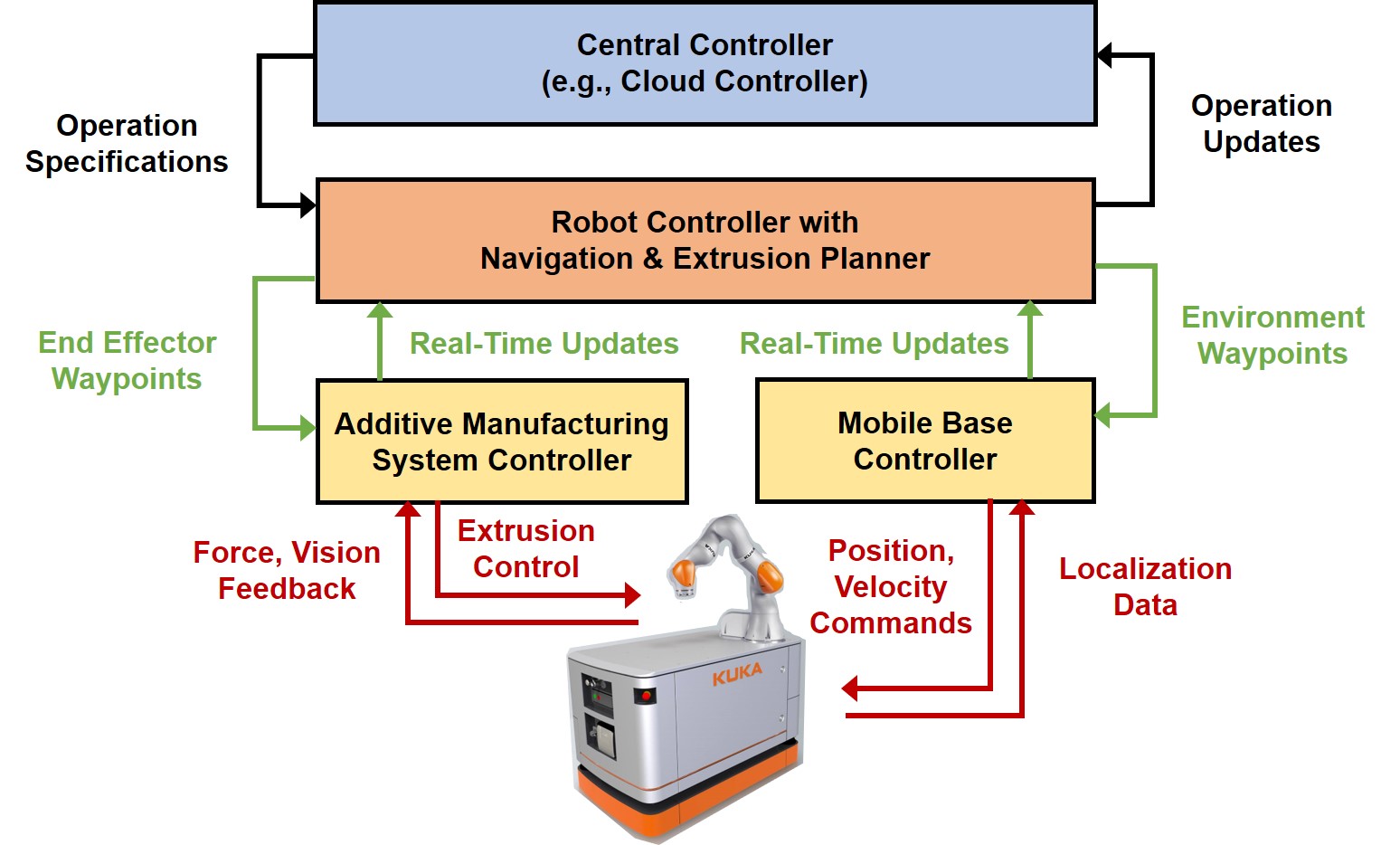}
\caption{Overview of the proposed framework.}
\label{fig:overview}
\end{figure}

\section{Mobile Additive Manufacturing Robot System}
\label{sec:methods}
The ability to seamlessly integrate different modalities of AM processing (e.g., pellet/wire/auger fed extrusion, inspection, pick-and-place) has been successfully demonstrated for metals, polymers, and concrete for a wide range of applications~\cite{coronel_increasing_2017,urhal_robot_2019}. However, there is a critical need to systematically assimilate the production opportunities and limitations of robots-enabled AM system across different ``layers'' of: product design~\cite{sauerwein_exploring_2019}, processing conditions~\cite{woern_fused_2018}, product mix as a function of resources~\cite{cruz_sanchez_plastic_2020}, and production demand and scheduling~\cite{strong_hybrid_2019}. In this research, the team presents a high-level description and a case-study demonstration of this system (shown in Figure~\ref{fig:overview}) that encompasses a mobile robot-assisted AM system which takes into account process control (in this case, polymer filament extrusion, but could be extended to other AM processes), mobile AM control (e.g., relative position to feedstock, point of need manufacturing, communication and monitoring), system level planning control (e.g., environment vs. end effector motion control), and cloud-based central controller (in this case, demonstrated with a single mobile base, but could be extended to co-mobile robots).

\subsection{Control of Mobile AM Robotic System}
\label{sec:control}

The proposed mobile additive manufacturing robotic system contains two primary sub-components: an additive manufacturing system and a mobile platform that moves the AM system and navigates the manufacturing environment.
The objective of the mobile additive manufacturing robotic system is to produce a part that meets product specifications, deadline requirements, and environment constraints.
Coordination between the two components will be necessary as the environment may require changes to AM process parameters or the AM process may require changes to mobile system parameters.
For example, the mobile system may be entering a space with an uneven or bumpy floor, requiring the AM system to slow down the extrusion process; or critical sections of the AM process may require the mobile system to slow down or stop to ensure that vibrations from the mobile base do not cause a failure for the print of the part.

Mobile manipulators have been expanding the capabilities of robotics systems. Mobility and manipulation capabilities allow mobile manipulators to be used in many scenarios, such as handling and assembling products in factories ~\cite{yang2019collaborative, ansari20225g} and maintenance and production in outdoor environments ~\cite{liu2023autonomous, rocha2021rosi}. Robotic AM control is enabling the fabrication of complex structures with high precision and reducing both support material waste manufacturing time ~\cite{urhal2019robot, li2018novel}. However, there is a shortage of control systems that effectively combine both mobile platform, robot arm, extrusion nozzle, and the high precision AM processes in dynamic environment.

We propose the use of a hierarchical control structure to ensure effective cooperation between the system components.
As shown in Figure~\ref{fig:overview}, the mobile system and the AM system will have individual controllers.
The AM system controller is responsible for determining parameters associated with the AM process, e.g., extrusion speed, and extruder movement.
The mobile base controller is responsible for ensuring following of a planned path in the manufacturing environment.
The robot controller will take in real-time information from the mobile base and the AM system to determine the end effector waypoints for the AM system and the environment waypoints for the mobile base.
This controller will use models of the extrusion process and the manufacturing environment to adapt the end effector waypoints and environment waypoints to ensure that the product follows manufacturing specifications and safely navigates the manufacturing environment.

In Section~\ref{sec:case1_mobility}, we analyze the effect that mobility has on the printing process.
In the case study, a robotic system moves a commercial 3D printer on the shop floor.
In the example, there is limited information sharing between the mobile platform and the additive manufacturing system is provided.
This case study showcases the need for coupling the mobile manipulator and the additive manufacturing process through a robot controller to ensure that the produced parts meet desired tolerances.

\subsection{System-level scheduling for Mobile AM}
\label{sec:scheduling}

The ``manufacture and deliver'' scheme for AM, initially proposed by Amazon and documented in a filed national patent in 2015 by Amazon Technologies, has garnered increasing research attention \cite{apsley2018vendor}. Referred to as ``mobile AM'' by~\cite{diri_kenger_integrated_2021}, this scheme introduces an integrated production-inventory-transportation (PIT) structure. In this innovative approach, the stages of production (3D printing), inventory, and transportation are seamlessly combined into a unified process. \cite{CUI2023108756} Investigates the delivery cost of products produced by mobile 3D printers in transit and strives to optimize operations from an economic standpoint. Despite the limited commercialization of mobile 3D printing on a large scale due to quality concerns, these exploratory studies are pivotal in advancing AM implementations. To the best of our knowledge, there are no studies that attempt to optimize the mobile AM scheduling considering vibration and battery charging.  For practical application, we propose a system-level scheduling for mobile AM considering movement vibration and robot battery charging.

The scheduling of mobile additive manufacturing systems presents a multifaceted challenge, compounded by the integration of assembly process constraints, traffic management, and battery charging considerations. Coordinating the additive manufacturing process with subsequent assembly steps introduces complexities that demand meticulous planning. Technical problems arise in aligning the temporal requirements of 3D printing with the downstream assembly tasks, ensuring seamless workflow integration. Simultaneously, traffic management poses a challenge in optimizing the movement of mobile manufacturing units within a given space, mitigating congestion, and ensuring timely progression through the manufacturing pipeline. Moreover, the coordination of battery charging schedules adds an additional layer of intricacy, as the system must intelligently balance printing, assembly, and recharging cycles. Addressing these technical hurdles necessitates the development of sophisticated scheduling algorithms capable of harmonizing diverse processes, considering resource constraints, and dynamically adapting to real-time changes in the manufacturing environment. Successfully navigating these challenges is pivotal for the efficient and effective deployment of mobile additive manufacturing systems in dynamic and resource-constrained settings.

In Section~\ref{sec:case2_system}, we provide results from the initial development of the autonomous manufacturing system consisting of a set of AM machines, AMRs, an input/output (I/O) station, and an assembly area. The AM machine produces parts with different processing times for each part ($PT$), assuming they are attached to each AMR. AMR drops the produced part in the assembly area and one AMR is allocated to assemble the parts into a product with an assembly time ($AT$). After assembling, an AMR picks up the product and returns it to the I/O station.

\section{Case Studies}
\label{sec:results}

Two case studies showcase some of the potential benefits of integrating the proposed mobile additive manufacturing robot system in the shop floor.
The first case study overviews how mobility can affect the additive manufacturing process and showcases the need for developing a platform that enables cooperation between the mobile base and the additive manufacturing system.
The second case study shows how the integration of the proposed system can improve manufacturing system throughput in manufacturing job shop.


\subsection{Effect of Mobility of AM Process}
\label{sec:case1_mobility}

\begin{figure}[t]
\centering
\includegraphics[width=0.6\columnwidth]{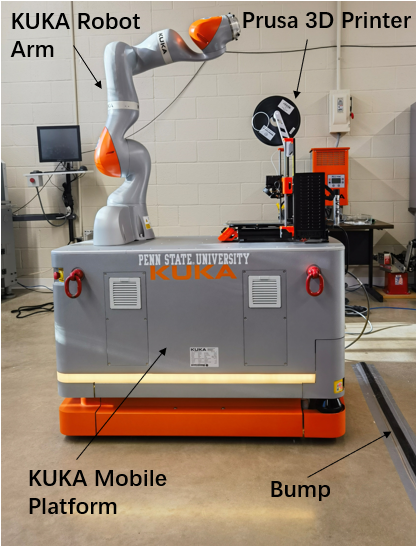}
\caption{The experimental mobile 3D printing platform consisting of a mobile robot and a 3D printer. The robot arm will be used to extend the proposed approach in future case studies, as described in Section~\ref{sec:conclusion}.}
\label{fig:testplatform}
\end{figure}

We have developed a research platform to determine the effects of mobility on a Fused Deposition Modeling (FDM) machine. The research platform uses a KUKA KMR iiwa mobile robot (KMR) 
\cite{KUKAWeb}. The KMR is controlled via a Sunrise.OS operating system, which enables operators to operate it through the Smartpad, a handheld control panel designed for the KMR.
The goal of the AM system is to print an experimental part with a height of 40mm, a width of 20mm, and a depth of 2mm. The precision tolerance for the products must not exceed greater than ±0.05mm.  The test platform is shown in Figure~\ref{fig:testplatform}. To streamline the printing and transportation processes and assess the viability of AMR, we conducted a trial run by deploying a commercial Prusa 3D printer on a KMR platform. The commercial 3D printer uses an extrusion-based additive manufacturing (AM) process named FDM 3D printing. This FDM 3D printing is widely used for quickly producing parts and objects. After performing additional calibrations, the precision tolerance can be minimized to as little as ±0.05mm on all axes when the  Prusa printer prints in a stationary environment~\cite{PrusaWeb}.

\begin{figure}[t]
\centering
\includegraphics[width=1\columnwidth]{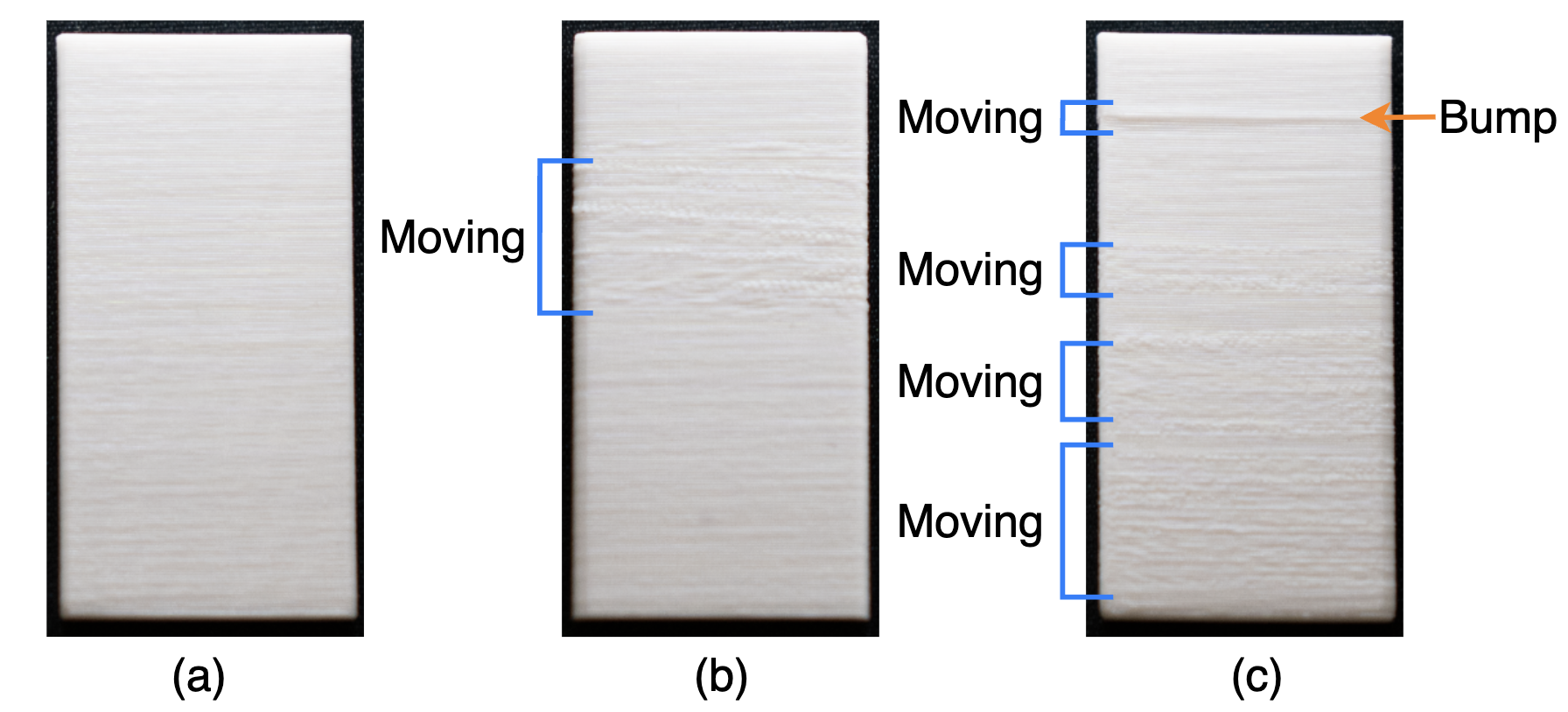}
\caption{Annotated images of the three printed parts for three different experiments: (a) a stationary print; (b) moving forward during a print; (c) moving forward and backward during a print several times, including over a bump during the last move. Note that the first layer is at the bottom of these images and the print direction was upward.}
\label{fig:moving3}
\end{figure}

\begin{figure}[t]
\centering
\includegraphics[width=0.7\columnwidth]{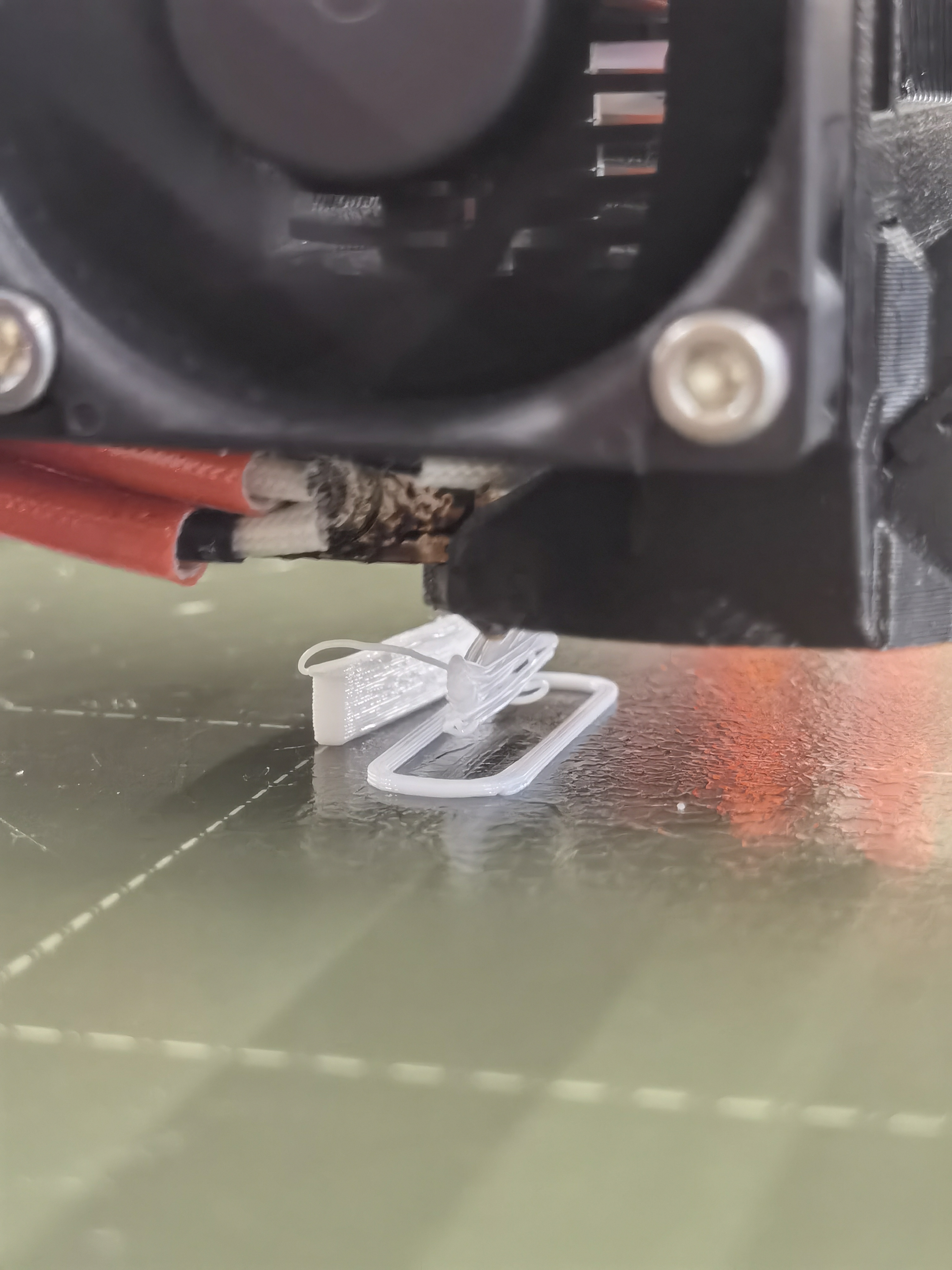}
\caption{The printed object detaching due to vibrations caused by the mobile platform going over the bump.}
\label{fig:moving4}
\end{figure}

In three test experiments, we use the Smartpad to control the movements of the KMR. In the first experiment, we tested the AM process when the system is stationary and measured the part dimensions using a digital caliper. The printed part, displayed in Figure~\ref{fig:moving3}a, exhibits a smooth and uniform surface layer texture and has dimensions of 39.96mm x 19.97mm x 2.01mm. These dimensions fit the tolerance requirements. In the second experiment, we tested the AM process while the KMR moving back and forth in a straight line. The part printed during this experiment is shown in Figure~\ref{fig:moving3}b and exhibits slight, but noticeable, irregularities in the surface layer texture. 
In addition, the final dimensions of the printed part are 40.09mm x 20.06mm x 2.05mm and do not meet the product specifications of ±0.05mm for the printed part dimensions.
In the third experiment, we tested the AM process moving back and forth on flat ground and through a 0.5-inch-high bump. The part printed during this experiment is shown in Figure~\ref{fig:moving3}c exhibits surface layer texture with noticeable defects. 
These defects are not acceptable since a part that has dimensions of 40.06mm by 20.08mm by 2.16mm is manufacturing, which exceeds the required tolerance of ±0.05mm.
Compared to second experiment, the defects in the third experiment are exacerbated as the KMR moves across the bump.
In addition, during the third experiment, the vibrations generated as the test print platform moves over the bump sometimes caused the printed object to detach from the print heat bed and result in a failure of the printing process, as shown in Figure~\ref{fig:moving4}.
Still, the initial findings indicate the viability of 3D printers being able to print while on the move due to the self-suspension capabilities of both the robot and the 3D printer.
Consequently, when the KMR operates with low acceleration and minimal vertical vibration, the surface of the printed object displays no noticeable defects within the acceptable range.

Future work will focus on extending the existing experimental testbed with an industrial robot arm (Kuka LBR iiwa). Combining the extrusion nozzle with the robotics arm can allow for the development of synchronized movements with high precision. Adaptive control algorithms to coordinate the robotic arms and mobile base could improve the quality of the printed part. These algorithms will incorporate methods to account for factors such as terrain variability, dynamic changes, and emergency situations. For example, developing a more advanced suspension system, or temporarily pausing the print until the KMR reaches a flat surface can help remove unwanted defects in the part. Moreover, improving the adhesion between the printed object and the heated bed would be an effective strategy for preventing detachment during motion. Finally, the calibration and monitoring systems should also be integrated to make runtime adjustments. We will also measure surface roughness in future case studies to quantify how mobility affects the quality of the printed part.


\subsection{Mobile AM in a Flexible Manufacturing System}
\label{sec:case2_system}

To validate the proposed approach's productivity performance in a manufacturing system, we develop a simulation model and compare it with the benchmark approach similar to the one in~\cite{ju2016selective,ju2014bernoulli}. We consider 2 AMRs and 2 AM machines in a system assuming unlimited buffer and unlimited automatic material supply. Each AM is assumed to be a dedicated machine producing two different parts for assembly, each with different production times. The AMRs should transport the produced parts to an assembly area, and a robot assembles the product by combining the two parts. The assembly time of a product follows a uniform distribution (min, max) = $U(10, 20)$. The completed product is transported from the assembly area to the input/output (I/O) station. In the proposed approach, an AM machine is coupled to each AMR, whereas the existing approach assumes that the AM machines are located in a machining area. The distance between the machining area, assembly area, and I/O station are assumed to be equivalent. AMR vibration and battery charging constraints are not considered in this experiment.  Table~\ref{table:1} summarizes the experimental manufacturing environment.

\begin{table}
\caption{Summary of experimental manufacturing environments.}
\begin{tabular}{p{4.8cm} p{3.5cm}}
    \hline
     Parameter & Values \\
     \hline
     Number of AMR & 2 \\
     Number of AM machine & 2 \\ 
     AM machine \#1 $PT$ (unit time) & $U(20,40)$ \\
     AM machine \#2 $PT$ (unit time) & $U(30,70)$, $U(40,80)$ \\
     AMR moving distance (unit distance) & 15, 30, 45 \\
     Assembly time (unit time) & $U(10,20)$ \\
     Simulation replication & 20 \\
     \hline
\end{tabular}

\label{table:1}
\end{table}

\begin{figure*}[t]
\centering
\includegraphics[width=1.8\columnwidth]{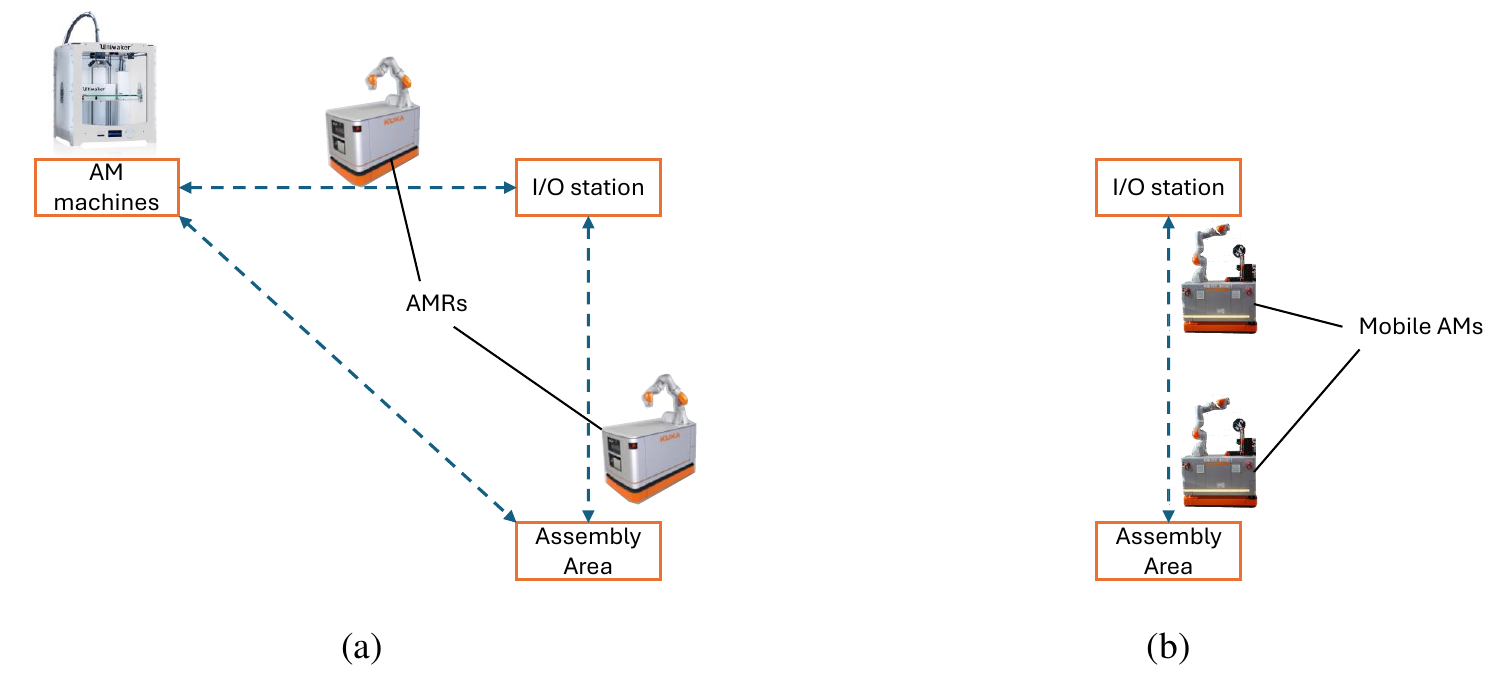}
\caption{Examples of manufacturing system simulation: (a) the existing approach; (b) the proposed approach.}
\label{fig:7}
\end{figure*}

We use ARENA software (version 16.20, Rockwell Automation, Milwaukee, Wisconsin) on a Windows 11 platform (Intel(R) Core(TM) i5-13600 CPU @ 2.70GHz, 32 GB memory, x64-based processor) to simulate the manufacturing scenarios as shown in Figure~\ref{fig:7}. To assess performance in different environments, we compare them across six manufacturing scenarios. We create these scenarios through a 3 $\times$ 2 full factorial design, which includes three AMR moving distances and two bottleneck AM machine processing times as shown in Table.~\ref{table:2}. We execute 20 random instances for each scenario, with each simulation run lasting one day (24 hours), including a warm-up time of 3 hours.

\begin{table}
\caption{Manufacturing scenarios.}
\begin{tabular}{l | l l}
    \hline
     Scenario & AMR moving distance & AM machine \#2 $PT$ \\
     \hline
     1 & 15 & $U(30,70)$ \\
     2 &  & $U(40,80)$ \\
     3 & 30 & $U(30,70)$ \\
     4 &  & $U(40,80)$ \\
     5 & 45 & $U(30,70)$ \\
     6 &  & $U(40,80)$ \\
     \hline
\end{tabular}

\label{table:2}
\end{table}

\begin{figure}[t]
\centering
\includegraphics[width=1\columnwidth]{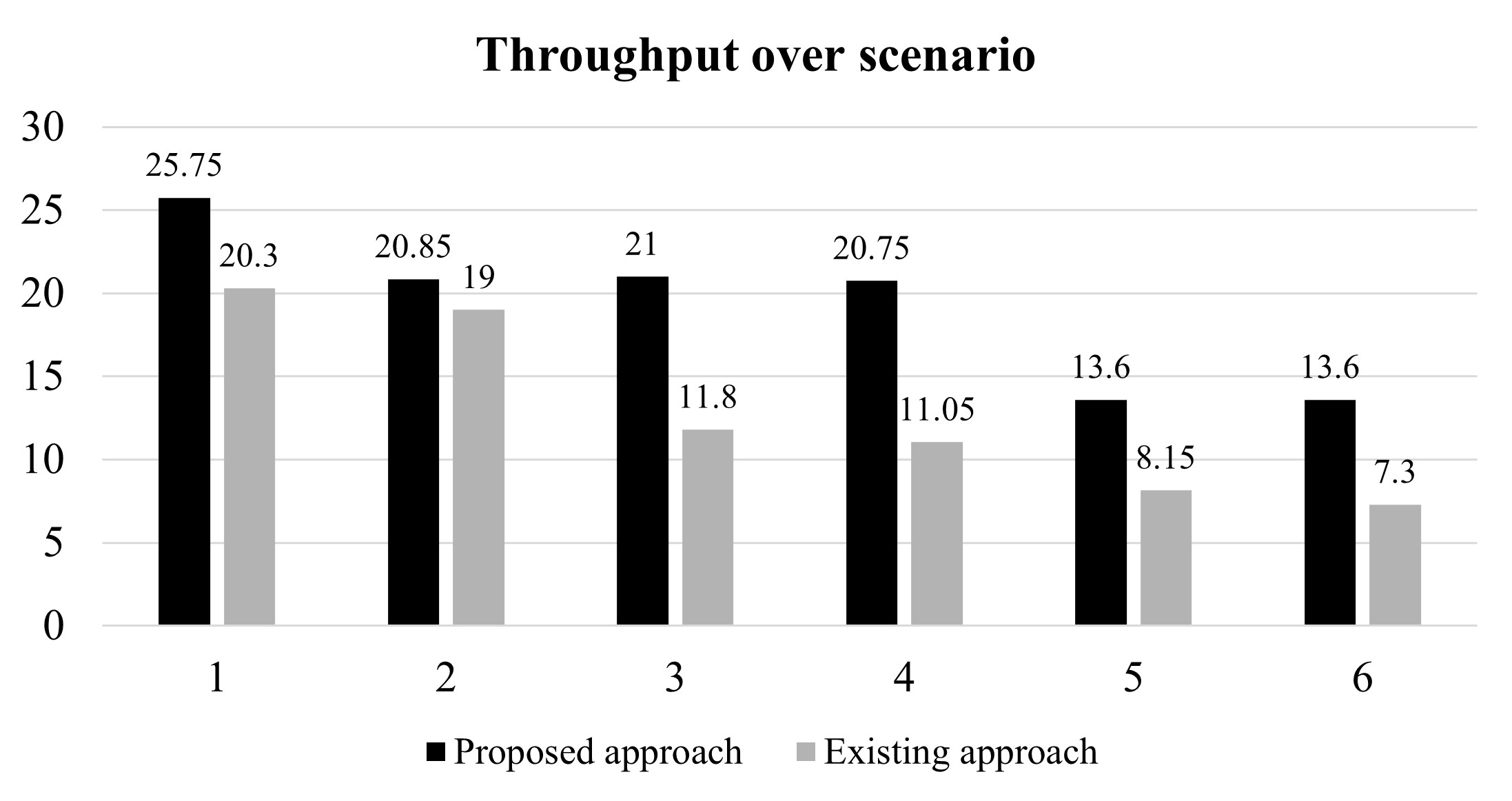}
\caption{Productivity performance over manufacturing scenario.}
\label{fig:8}
\end{figure}

Figure~\ref{fig:8} illustrates the average throughput of each approach for the finished product across scenarios. The proposed approach consistently outperforms the existing method in terms of throughput across all scenarios. This is attributed to the fact that the AMR travel distance in the existing approach is larger than that in the proposed approach, mainly due to the additional travel distance to the separate processing area. As the processing time for AM machine \#2 increases and the AMR travel distance extends, there is a noticeable decreasing trend in throughput. The throughput of the proposed approach in scenario 2-4 remains consistent because when the processing time for AM machine \#2 follows a uniform distribution $U(40,80)$, it exceeds the travel time of the AMR, and when the AMR moving distance is 30, the AMR travel time aligns with the AM machine \#2 processing time. In the existing approach, the AMR travel time becomes the bottleneck in completion time across all scenarios. This indicates that additional analysis is required to develop and validate decision making heuristics that can concurrently control the process level- (AM), product level- (design), unit level- (mobile robotic AM), system level- (co-mobile robotic AMs), and enterprise level- (cloud computing based resource planning). 

Future research on simulation experiments will take into account practical manufacturing systems and various constraints. In the above simulation, the number of AMRs and AM machines is limited to a small-sized toy example. A more extensive consideration of AMRs, AM machines, and part types is necessary to address practical manufacturing systems. The AM machines should be versatile equipment capable of producing various kinds of parts, and additional constraints should be introduced, such as buffer size and the method of feeding materials. All parameters, including production time, AMR travel distance, and assembly time, will be redefined based on real industry references. The system will incorporate a charging station for AMRs, and battery charging constraints will be considered in the scheduling process. Furthermore, we plan to optimize the scheduling of a mobile AM manufacturing system with consideration of product quality standards and productivity. The proposed approach's performance will be validated through simulation experiments.

\section{Conclusion}
\label{sec:conclusion}

Manufacturers are often looking to improve the flexibility and adaptability of their systems to meet rapidly changing customer requirements.
To meet both the customizability requirements and tight deadline constraints, we propose a mobile additive manufacturing systems that combines Additive Manufacturing (AM) and Autonomous Mobile Robotics (AMR) for ``manufacture and deliver'' capabilities. At the process level, we emphasize the development of an extrusion attachment, leveraging existing robotic AM technology for rapid part production. The system's adaptability at the machine level accommodates dynamic changes in the manufacturing environment, optimizing the movement of both the extruder and the AMR. 

Our research contributes to the industrial AM landscape, presenting a novel perspective on mobile additive manufacturing for smart manufacturing systems. By combining manufacturing and delivery capabilities, our framework can improve the scheduling and planning processes and introducing improved levels of customization and efficiency.
The proposed approach may work well for larger factory floors, where the movement time significantly affects the flow time for a part.
In addition, it may be useful for emergency scenarios where every second or minute matters for the manufacturer, e.g., a maintenance task that shuts down an entire line and requires the printing of a specific component or part.
In addition, while this research uses polymer additive manufacturing in conjunction with a mobile base, results from this work can be extended to other production scenarios such as concrete printing, printing during transit, and in-space manufacturing.

Future work will be directed to further investigating the Robotic AM process and system integration to determine the full capabilities of mobile and additive manufacturing for smart manufacturing systems.
Specifically, the next steps of this research will be the creation of a multi-level intelligent decision making model for Robotic AM manufacturing systems for on-demand, on-site, and cloud-computing controlled manufacturing of customized products. 

\section*{Acknowledgment}
The work is in part supported by NSF under grant CMMI-1922739. 

\balance

\end{document}